\documentclass[journal]{IEEEtai}

\usepackage[colorlinks,urlcolor=black,linkcolor=black,citecolor=black]{hyperref}
\usepackage{color,array}

\usepackage{graphicx}

\usepackage{multirow}%
\usepackage{amsmath,amssymb,amsfonts}%
\usepackage{amsthm}%
\usepackage{mathrsfs}%
\usepackage{xcolor}%
\usepackage{textcomp}%
\usepackage{manyfoot}%
\usepackage{booktabs}%
\usepackage{algorithm}%
\usepackage{algorithmicx}%
\usepackage{algpseudocode}%
\usepackage{listings}%
\usepackage{subcaption} %
\usepackage{tabularx}  

\begin{document}

\title{Improved Alignment of Modalities in Large Vision Language Models}

\author{
    \IEEEauthorblockN{Kartik Jangra, Aman Kumar Singh, Yashwani Mann, Geetanjali Rathee} \\
    \IEEEauthorblockA{
        \textit{Netaji Subhas University of Technology}\\
        \href{mailto:kartik-ug21@nsut.ac.in}{kartik-ug21@nsut.ac.in}, 
        \href{mailto:aman.singh.ug21@nsut.ac.in}{aman.singh.ug21@nsut.ac.in}, 
        \href{mailto:yashwani.ug21@nsut.ac.in}{yashwani.ug21@nsut.ac.in}, 
        \href{mailto:geetanjali.rathee123@gmail.com}{geetanjali.rathee123@gmail.com}
    }
}

\maketitle

\begin{abstract}
Recent advancements in vision-language models have achieved remarkable results in making language models understand vision inputs. However, a unified approach to align these models across diverse tasks such as image captioning and visual question answering remains a challenge. Existing methods either require very big language models or very big datasets which is not efficient in utilizing existing models. This paper addresses this gap and devises a training strategy of auto-regressive vision-language models, to unify vision-language tasks like image-captioning and visual question answering. We propose four training stages for aligning the vision model with the language model, in other words, the language model
is given an ability to process visual inputs. We also devise different attention masks for training transformer-based language models that improve the quality of visual features. Further, we introduce some findings, 1) the attention mask
should not be applied on visual inputs, 2) the Language model converges faster on
AI-generated data, 3) More work should be done in the alignment stage during the pre-training of the model, 4) the model can easily adapt to any downstream tasks like visual question answering on healthcare datasets like PathVQA. After training the model for one epoch for all the stages, it outperforms large models like
VILA-13 billion models on common benchmarks like CIDEr scores on COCO and Flickr30k datasets and achieves very close scores to GIT-2 on the same dataset
despite being a much smaller model trained on a much smaller dataset. All of the training is done using best practices available like multi-GPU parallel training, lower-precision training with 16-bit float numbers, faster attention (SDPA), and
gradient accumulation, and completed the training within 12 hours.
\end{abstract}

\begin{IEEEImpStatement}
In the race towards Artificial General Intelligence, there is a need for a modal with multiple capabilities, known as multimodal models. The vision-language models have capabilities of both image and language. There have been many attempts to create a good vision-language model, but it always demands great computation and training costs as these models are very big, they have billions of parameters, and it requires a large amount of computational cost even for inference. Training and using such big models leaves a large amount of carbon-footprint and affects the environment in a negative way. This paper explores the reason for their big sizes, and devises a training recipe and a vision-language model to mitigate the limitation of these models' high-computation requirements. This work proposed a 900-million parameters model than outperformed much larger models and showed comparable performance on models of similar size but trained on 100 times larger dataset. Inference of this model can be done even on CPU and could be utilized by businesses with limited funding and in healthcare for general diagnosis as proposed model showed promising result when trained on medical dataset. In healthcare, vision-language models assist in generating detailed medical image reports, diagnosing conditions from visual data, and supporting patient-doctor communication. They improve accessibility by translating visual content into text for better understanding. For general use VLMs enhance multimedia search, automate image captioning, and enable seamless interactions in applications like virtual assistants and educational tools. 
\end{IEEEImpStatement}


\section{Introduction}

\IEEEPARstart{T}{ransformers}-based Large Language models (LLMs) \cite{Vaswani2017AttentionIA} have shown remarkable performance on natural language tasks. Following the same architecture  Vision Transformers \cite{Dosovitskiy2020AnII} came up and showed comparable performance on computer vision tasks to deep CNN models. Since then there have been many successful attempts in augmentation of LLMs to support visual inputs. This work focuses on pre-training of a vision language model for better alignment of both modalities which doesn't degrade the individual capabilities of both language and vision transformers.  With the advancement of conversational AI models, various industries are adopting rapid change.

Tremendous advances have been made in transformer-based multimodal pre-training starting from  CLIP \cite{Radford2021LearningTV}, SimVLM \cite{Wang2021SimVLMSV}, Pali-Gemma \cite{Beyer2024PaliGemmaAV}, LLaVA-series \cite{Liu2023VisualIT, Li2024LLaVAOneVisionEV, Li2024LLaVANeXTInterleaveTM}. Some of them train both the models from scratch like GIT-2 \cite{Wang2022GITAG} and some models use pre-trained vision and language transformers like LLaVa series and Pali-Gemma. Pre-training the transformer-based models is a very expensive task, so the most cost-effective solution is to use existing good vision and language models and align them, or make a common embedding space for visual inputs and text inputs.
In this work, we explore different methods used for alignment and pre-training the vision-language model which can later be fine-tuned on various downstream tasks. We followed the LLaVa one-vision style alignment but with much fewer vision tokens and different attention masks. We discover several findings on what type of data helps the pre-trained model converge faster for the generation task as discussed in section \ref{sec4}.

\subsection{Common components used in designing the vision language models are:}

1) \textbf{Vision Encoder}: The features from the image are extracted using vision encoders, it can either be a deep CNN model like  Vgg16 \cite{Simonyan2014VeryDC}, ResNet50 \cite{He2015DeepRL}, or a newly emerged Vision Transformer(ViT) \cite{Dosovitskiy2020AnII} models. 
For ResNet50, the feature map is of dimension 7*7*2048, it can either be reshaped into 49*2048, or pooled to 7*2048. In ViT, the image itself is divided into a fixed number of patches of fixed size. Each patch can be represented using a vector of dimension $d_{\text{im}}$, (for example 224*224 image can be divided into 14*14 patches constituting a total of 256 patches), so an image can be considered a sequence of 256 features. Since it is very intuitive to see the image as a sequence of features in ViT and is also a transformer-based model that follows similar architecture to the Language Model, it has become a popular choice as a Vision Encoder for the vision language model. The most popular Vision Transformers are CLIP-ViT of SigLIP-ViT because both have been pre-trained for large amounts of Image-Text pair data. \\
2) \textbf{Projector}: The $N_{\text{im}}$ is number of features in image with dimension $d_{\text{im}}$
 and $N_{\text{txt}}$ is number of features in text with dimension $d_{\text{txt}}$. The language model expects the features to be in $d_{\text{txt}}$, so the image features must be projected to an embedding space of dimension $d_{\text{txt}}$. The projector can be a simple linear layer, an MLP layer, or a transformer block.\\
3) \textbf{Language Model}: This is a decoder-style transformer model. This is where the generation of text takes place. The model takes projected image features and text features as inputs and uses cross-entropy loss for calculating loss for next-token prediction(Causal Language Modeling). This model can be a pre-trained model or a model without any pre-training. 

Common Datasets used for training.

1) \textbf{Pre-training}: The very first pre-training stage is done on data with image-text pair, popular open source data are  LAION-5B \cite{Schuhmann2022LAION5BAO},  COYO-700M \cite{kakaobrain2022coyo-700m}, and interleaved image-text data like  MMC4 \cite{Zhu2023MultimodalCA}.

2) \textbf{Instruction Fine-tuning}: This is the fine-tuning stage of the model, Various instruction-based open source datasets are available like VQAv2, GQA, COCO-captions, LLaVA-Next dataset \cite{Li2024LLaVANeXTInterleaveTM}, LLaVA-Instruct \cite{Liu2023VisualIT}.

\subsection{Research Gaps}

Although the vision-language model is a hot topic among researchers, there is a lack of understanding of how language models can pay attention to image patches. For example, LLaVA series\cite{Liu2023VisualIT, Li2024LLaVAOneVisionEV} models split the images into a number of images, why does this work better without splitting the image? The answer to this question is an attention mask. After splitting the image, each and every patch from those split images can pay attention to every other patch
even with causal masking. This behavior can be replicated with a sequence-to-sequence mask without any image splitting resulting in a much lesser number of image tokens than that of LLaVA series. 

\begin{figure}[t]
    \centering
    \includegraphics[width=0.9\linewidth]{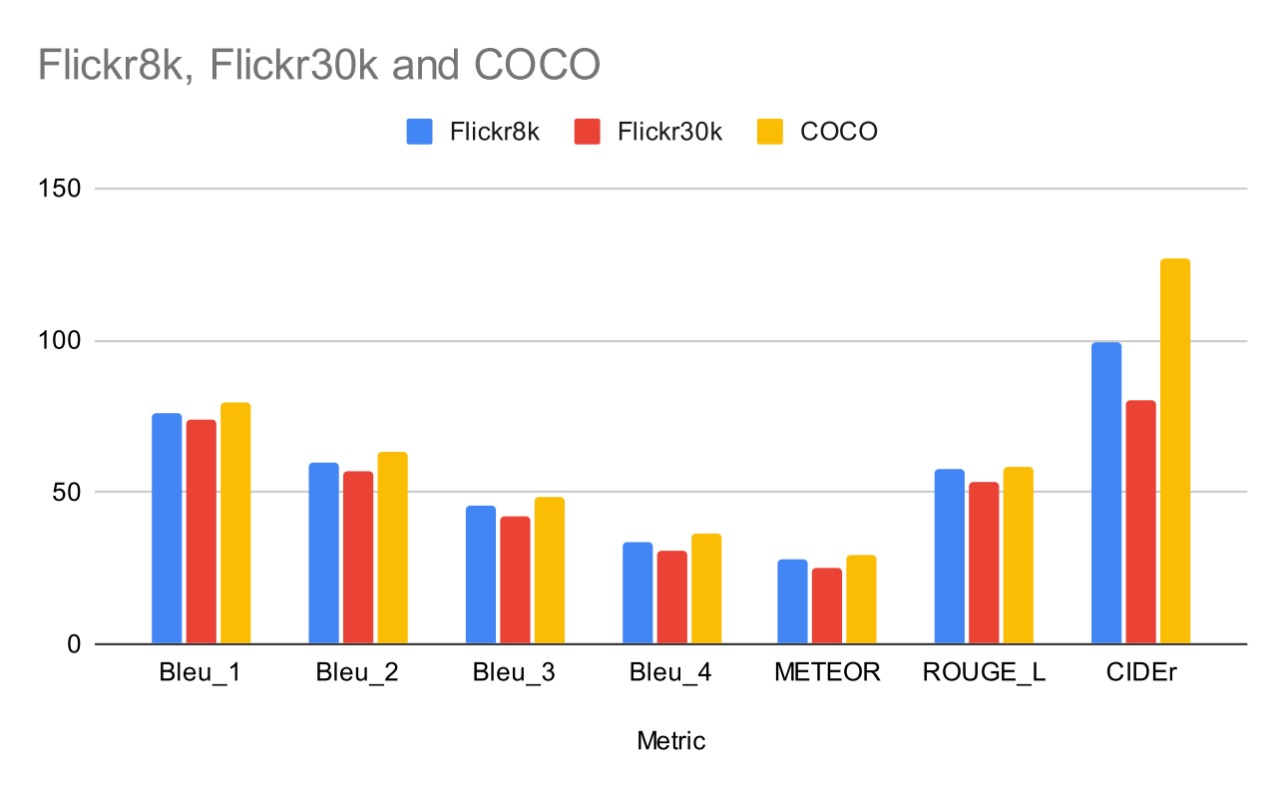}
    \caption{Model performance on different datasets.}
    \label{fig:metric_compare}
\end{figure}

\subsection{Contribution of paper}
The main contribution and findings of this work are:

1) Alignment of the models is very crucial before pre-training the foundational vision and language models, alignment can be done in two different ways, the first one is to train only the projector module, and the second one is to train both the vision transformer and projector.
2) For two datasets, with the same images but one has human-generated captions and the other one has AI-generated captions, the model converges faster with AI-generated captions.
3) Dividing the images into grids of images as done in the LLaVa series can be replaced simply by unmasking the image during the training stages.  
4) The cross-entropy loss for two input statements human-generated and AI-generated on a similar topic, is lesser for AI-generated text.

We introduced a step-wise training method to design a Large Vision-Language Model, that achieves comparable performance to SOTA methods with a much smaller vision model, much smaller visual input tokens, and a much smaller language model. There is a memory issue in large language models, so it is not efficient to run on edge devices with limited computational power, Therefore, it is essential to have small - Large Multimodal models which can run efficiently on edge devices. The model can be applied to solve various problems such as in Healthcare, and virtual assistants. Small businesses can afford to run these small models with comparable performance to big models which requires a large amount of computational resources which are very expensive to afford. For example, Llama \cite{Touvron2023LLaMAOA} 3-400 billion parameter model would require 1.6 terabytes for 32-bit precision, 0.8 terabytes for 16-bit precision, and 0.4 terabytes for INT8 precision.
To solve this problem we followed a  LLaVa-Next style, a training method with much less image sequence and a smaller vision transformer, and explored new ways to align the vision features into the text embedding space.

The remaining structure of the paper is organized as follows. The section discusses the existing models and techniques to improve knowledge performance. Section three details the proposed multi-model network to improve the alignment of modalities using different attention masks. Further, section four evaluates the proposed model on various tasks with fine-tuning (full-shot) and without fine-tuning (0-shot). We used Flickr8k, Flickr30k, and MS-COCO, for the model's general world knowledge performance on metrics like BLEU1, BLEU2, BLEU3, BLEU4, METEOR, ROUGE1, and CIDEr scores as shown in Figure .\ref{fig:metric_compare} and further analysis is discussed in section \ref{sec4}. For domain-specific tasks, we fine-tuned the model on healthcare datasets mainly PathVQA \cite{He2020PathVQA3Q}. Moreover, section \ref{sec5} concludes the paper along with future directions of the paper.

\section{Related Work}\label{sec2}

\begin{table*}[h]
\caption{Literature Survey }\label{tab:survey}
\begin{tabularx}{\textwidth}{|X|X|X|}
\toprule
\textbf{Authors} & \textbf{Method Used} & \textbf{Key Findings} \\
\midrule
\cite{Radford2021LearningTV} & Contrastive Pre-training & Images and text can share the same embedding space. \\ 
\hline
\cite{Wang2022GITAG} & Seq2Seq Language Modeling & Vision-language models can be trained auto-regressively. \\ 
\hline
\cite{Alayrac2022FlamingoAV} & Introduced a perceiver resampler as an image feature projector and connected image and text features using cross-attention while keeping the language model weights frozen. & Vision-language models can be adapted to various tasks like image captioning, visual question answering, and multi-image question answering. \\
\hline
\cite{Liu2023VisualIT, Li2024LLaVAOneVisionEV, Li2024LLaVANeXTInterleaveTM} & Generated synthetic text for existing open-source image-text datasets for visual-instruction fine-tuning. & The model performs better with synthetically generated text. \\
\hline
\cite{Li2023BLIP2BL} & Introduced a query-former for learning queries for image features and trained the model on image-text matching, contrastive loss, and generative language modeling. & BLIP-2 bridges image and language modalities using a two-stage trained Q-Former, achieving state-of-the-art results in tasks like VQA, image captioning, and image-text retrieval. \\
\hline
\cite{Beyer2024PaliGemmaAV} & Followed the same strategy as GIT-2, added instruction and image as prefix of the sequence. & Seq2Seq masking improves the quality of result in vision-language models.\\
\bottomrule
\end{tabularx}
\end{table*}

There has been a tremendous amount of work done in training vision-language models and still currently a hot topic among researchers. The transformer-based vision language models can be classified into two categories, one is  Auto-regressive models and the other one is fusion-based models.
Fusion-based models use techniques like cross-attention, between image features and text features for tasks like masked language modeling, masked Image modeling, and generative language modelling like  SimVLM\cite{Wang2021SimVLMSV}, Uniter\cite{Chen2019UNITERUI} and BLIP\cite{li2022blip}. They follow a standard encoder-decoder transformer model\cite{Vaswani2017AttentionIA}, where the Encoder is the vision encoder and the Decoder is the text decoder. This type of architecture is usually trained on various image-text objectives like masked-image modeling, masked-text modeling, image-text matching, image-text contrasive training, and language modeling. \\
The second one is auto-regressive models that use a decoder-only transformer for both image features and text features. It consists of \textbf{Vision Encoder}, which maps the images into sequence of features, a text \textbf{Tokenizer} that maps a sequence of text into sequence of features, a vector \textbf{Projector} that bridges the gap between the image features and text features, and a \textbf{decoder} only transformer(Large Language Model) that takes these vision and text features as a single sequence as inputs, where image features are prefix and text features is the suffix of the sequence, this way the decoder(LLM) can be trained auto-regressive way, models like like \cite{Wang2022GITAG, Liu2023VisualIT, Li2024LLaVAOneVisionEV, Beyer2024PaliGemmaAV} are trained using this recipe.

Table \ref{tab:survey} presents the literature survey of the paper.

\section{Proposed Method}\label{sec3}
This section details the proposed multi-model network and training methods by explaining the model architecture, attention masking process, dataset mixture, and training stages along with training methods. 

\subsection{Model Architecture} 

\begin{figure}[t]
    \centering
    \includegraphics[width=0.8\linewidth]{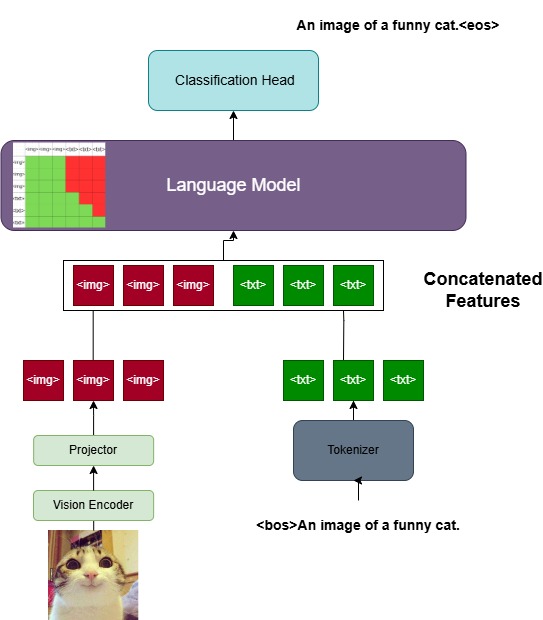}
    \caption{The model architecture, it consists of a vision encoder, a multi-modal projector and a language model.}
    \label{fig:architecture}
\end{figure}

The model has three components, Vision Encoder, Projector, and Language Model as shown in Fig \ref{fig:architecture}. For the vision encoder we use  SigLip ViT \cite{Zhai2023SigmoidLF} for its better performance than CLIP-ViT \cite{Radford2021LearningTV}, and it has already been pre-trained for image-text Contrastive Loss. The model takes inputs of the image of size 224X224 and divides the image into patches of size 14X14, each patch is independent of other patches, to produce 256 such patches. Each patch is passed through convolutional operation and then through a bidirectional transformer encoder to produce an embedding vector of dimension $d_{\text{V}}$. This way the image is represented by:
\[
\mathbf{V} = \left[ \mathbf{V}_1, \mathbf{V}_2, \dots, \mathbf{V}_{256} \right], \quad \mathbf{V}_i \in \mathbb{R}^{d_{\text{V}}}
\]
Where each $\mathbf{V}_i$ is the embedding vector corresponding to each patch, and $256 = 16 \times 16$ is the total number of patches.

The projector is a two-layer perceptron with a GeLU activation function, which maps the image representation $\mathbf{V}$ from the vision embedding dimension \( d_{\text{v}} \) to the text feature dimension \( d_{\text{L}} \), where \( d_{\text{L}} \) matches the dimension of the text embeddings. The projected embedding is represented by:
\[
\mathbf{P} = \left[ \mathbf{p}_1, \mathbf{p}_2, \dots, \mathbf{p}_{256} \right], \quad \mathbf{p}_i \in \mathbb{R}^{d_{\text{L}}}
\]
Where each $\mathbf{p}_i$ is the embedding vector corresponding to each image token.

For the language Model, we use Qwen-2.5 with a 0.5 billion parameter model. The text features are represented by:
\[
\mathbf{L} = \left[ \mathbf{L}_1, \mathbf{L}_2, \dots, \mathbf{L}_{seqLen} \right], \quad \mathbf{L}_i \in \mathbb{R}^{d_{\text{L}}}
\]
The text embedding of the language model and the projected embedding from the projector are concatenated to form a sequence. Input embedding of the model is  represented by:
\[
\mathbf{I} = \left[ \mathbf{p}_1, \mathbf{p}_2, \dots,\mathbf{p}_{256}, 
\mathbf{L}_{1} ,
\mathbf{L}_{2} ,
\dots, 
\mathbf{l}_{seqlen} \right], \quad \mathbf{L}_i, \mathbf{p}_i \in \mathbb{R}^{d_{\text{L}}}
\]
Where $\mathbf{I}_i$ from i = 0 to i = 255 is equal to $\mathbf{p}_i$ and all the next $\mathbf{I}_i$ are the embedding vector corresponding to each text token, and seqLen is the sequence length of the model.

\begin{table}[htbp]
    \centering
    \begin{subtable}[t]{0.45\textwidth}
        \centering
        \caption{Qwen2 Model Configuration}
        \begin{tabular}{|c|c|}
            \hline
            \textbf{Parameter} & \textbf{Value} \\
            \hline
            Hidden Size & 896 \\
            \hline
            Intermediate Size & 4864 \\
            \hline
            Number of Attention Heads & 14 \\
            \hline
            Number of Hidden Layers & 24 \\
            \hline
            Number of Key-Value Heads & 2 \\
            \hline
            Max Position Embeddings & 32768 \\
            \hline
            Vocab Size & 151936 \\
            \hline
            Torch Dtype & bfloat16 \\
            \hline
        \end{tabular}
        \label{tab:qwen2_config}
    \end{subtable}
    \hfill
    \begin{subtable}[t]{0.45\textwidth}
        \centering
        \caption{Siglip Vision Model Configuration}
        \begin{tabular}{|c|c|}
            \hline
            \textbf{Parameter} & \textbf{Value} \\
            \hline
            Hidden Size & 1152 \\
            \hline
            Patch Size & 14 \\
            \hline
            Number of Attention Heads & 16 \\
            \hline
            Number of Hidden Layers & 27 \\
            \hline
            Image Size & 224 \\
            \hline
        \end{tabular}
        \label{tab:siglip_config}
    \end{subtable}
\caption{Model Configuration}

\end{table}

This vector matrix $\mathbf{I}$ acts as the input embedding to the language model. The embedding corresponding to image embedding acts like a foreign language that the model has never seen before, so in order to not to degrade the language model learned parameters, freezing the language model in the initial stages of training is crucial. The task is to align the embedding of both the modalities and to be more specific, to bring $\mathbf{P}$ closer to $\mathbf{L}$. The model's Vision Encoder and Language model's specifications are given in Table \ref{tab:siglip_config} and Table \ref{tab:qwen2_config} repectively.

\subsection {Attention Mask}
The most important part of this work is studying the attention masking in transformer architecture. LLaVA models split the images into grids of images and show good performance with traditional causal masking. This is because after splitting and putting them into a sequence, most of the image patches can attend to other patches of the image which isn't possible without splitting the image. The same behavior of a patch paying attention to all other image patches can be replicated without splitting the image just by unmasking all the image attention scores between the image patches as shown in Fig. \ref{fig:attentionl}, and keeping the causal mask for text tokens, this way text tokens only attend to preceding tokens.
In Fig. \ref{fig:attentionl} and Fig. \ref{fig:full_unmaskl}, if the cell \(C_{ij}\) is green, it means the \(i\)-th token can attend to the \(j\)-th token, and if the cell \(C_{ij}\) is red, the \(i\)-th token cannot attend to the \(j\)-th token.

\subsection{Dataset Mixture}
\begin{table}[htbp]
\centering
\caption{Data mixture for all training stages}

\begin{tabular}{|c|c|}
\hline
\includegraphics[width=0.20\textwidth]{coco-recap.jpg} & 
\begin{tabular}{@{}l@{}}
\textbf{Stage-0 \& satge-1 | COCO-Recap-118k:}\\
\\ The image shows a meal served in a blue tray\\ with  compartments. In the top left compartment, \\ there is a slice of bread with a spread that appears\\ to be butter,  accompanied by a few almonds and a \\slice of what looks like a baked  potato or sweet \\ potato. The top right compartment contains a \\ variety of fruits, including what seems to be \\  pineapple, orange slices, and possibly a \\piece of melon.
\end{tabular} \\
\hline
\hline
\includegraphics[width=0.20\textwidth]{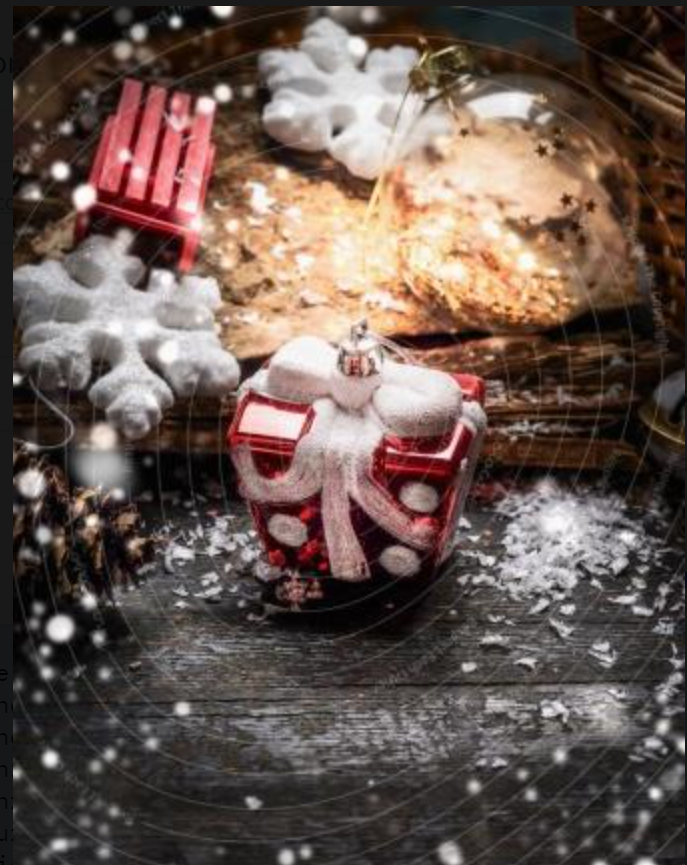} &
\begin{tabular}{@{}l@{}}
\textbf{Stage-2 | BLIP558k Recap:}\\
\\ The image depicts a festive and cozy scene, likely\\ set during the holiday season. In the center of the\\ image, there is a red and white gift box with a\\ ribbon, suggesting it is a Christmas present. The\\ box is placed on a wooden surface that appears to\\ be a table or a bench, which is covered with a\\ layer of what looks like snow, adding to the wintry\\ atmosphere.
\end{tabular} \\
\hline
\hline
\includegraphics[width=0.2\textwidth]{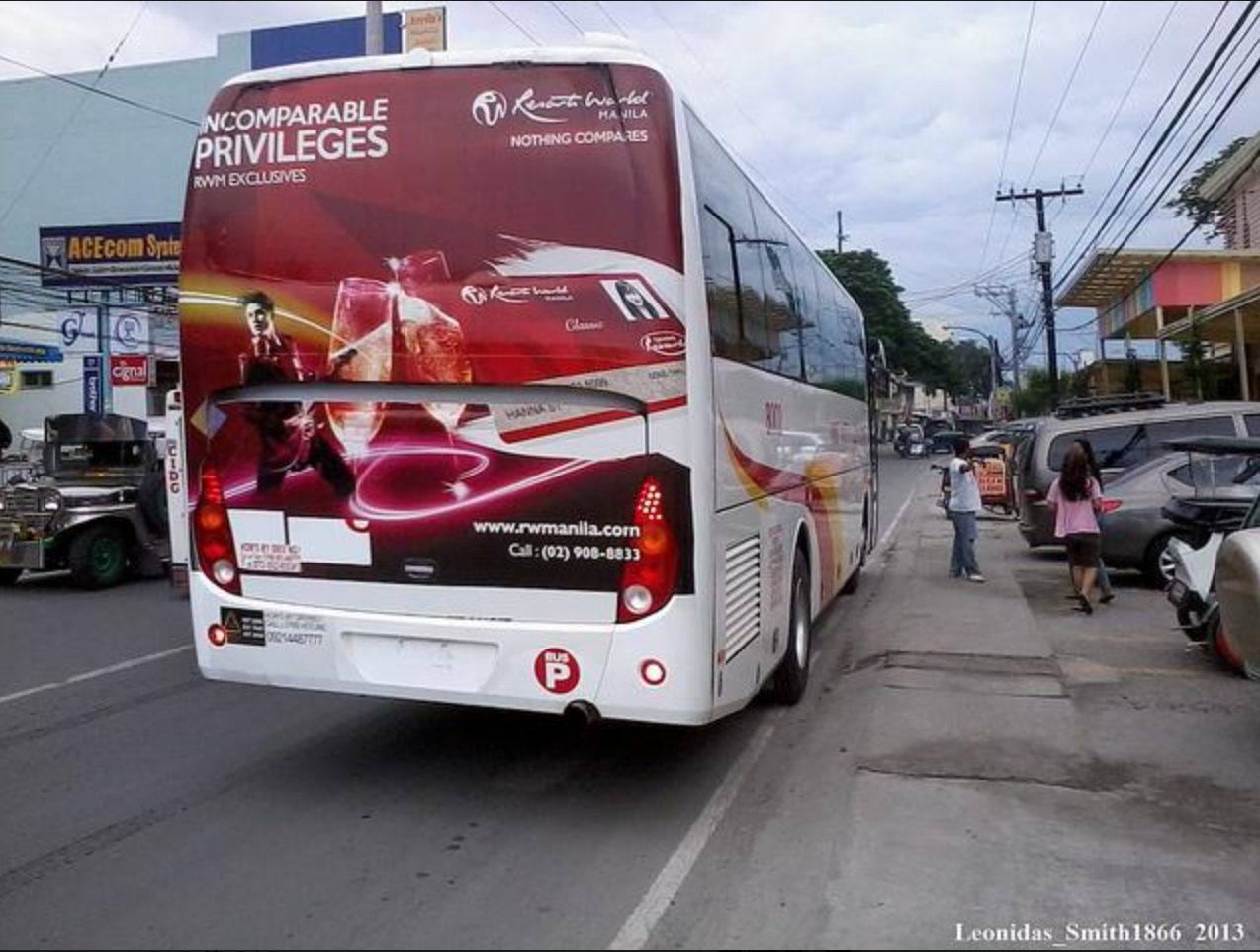} &
\begin{tabular}{@{}l@{}}

\textbf{Stage=3| LLaVA-Next-Data 790k:} 
\\ Que: What are the colors of the bus in the image?\\
Ans: The bus in the image is white and red.\\
\\Que: What feature can be seen on the back of the\\ bus?\\
Ans: The back of bus features an advertisement.\\
\\Que: Is the bus driving down the street or pulled\\ off to the side?\\
Ans: he bus is driving down the street, which is\\ crowded with people and other vehicles\\

\end{tabular} \\
\hline
\end{tabular}
\label{tab:data_mixture}
\end{table}

The model is trained on various publicly available datasets by LLaVA-Next authors. For stage-0 and stage-1, the model is trained on COCO-118k-Recap, for stage-2 BLIP558k-Recap and for stage-3 LLaVa-Next-790k Dataset is used. The detailed information about dataset is given in Table \ref{tab:data_mixture}.

\subsection{Training Stages}
We add an extra pre-training stage to the training strategy followed by LLaVA-Next and LLaVA-OneVision.

\begin{itemize}
\item \textbf{Stage-0}: Uses Bidirectional attention for better and faster alignment of vision features into text embedding space. We unmask all the attention scores so that image tokens, Fig. \ref{fig:full_unmaskl}, can attend to preceding as well as following tokens. The language model is kept frozen to keep it unaffected by the unmasking of the following tokens, only the Projector is trained. As the classification head is predicting the next word or token, and due to bidirectional attention model can pay attention to the next token and it may learn an identity function which may not lead to the alignment. To solve this problem we add noise in the input sequence by masking 20\% of the input tokens. The algorithm is discussed in Algorithm \ref{algo:1}.

\begin{algorithm}
\caption{Algorithm for Model Forward Pass for Stage-0}
\begin{algorithmic}[1]  
\State $V \gets \textsc{vision\_encoder}(image)$
\State $P \gets \textsc{projector}(V)$
\State $text \gets \textsc{add\_noise\_to\_input}(text)$
\State $T \gets \textsc{text\_embedding}(text)$
\State $L \gets \textsc{concatenate}(P, T)$
\State $attn\_mask \gets \textsc{unmask\_attention}(attn\_mask)$
\State $logits \gets \textsc{language\_model}(L, attn\_mask)$
\State $loss \gets \textsc{cross\_entropy\_loss}(logits, label\_text)$
\end{algorithmic}
\label{algo:1}
\end{algorithm}

\item \textbf{Stage-1}: Uses Unidirectional attention for the text tokens and bidirectional attention for the image tokens as shown in Fig. \ref{fig:attentionl}  In this stage of training we restrict the image patches to attend to the following text tokens by masking attention scores between image patches and text tokens. Here we don't mask the input because of causal masking for the text tokens, which means the text tokens cannot attend to the future tokens. We unmask the attention score between image patches, so there is no need for splitting of images as done in the LLaVA series. Refer to Algorithm \ref{algo:2}.   

\begin{figure}[htbp]
    \centering
    \begin{subfigure}[t]{0.3\textwidth}
        \centering
        \includegraphics[width=\linewidth]{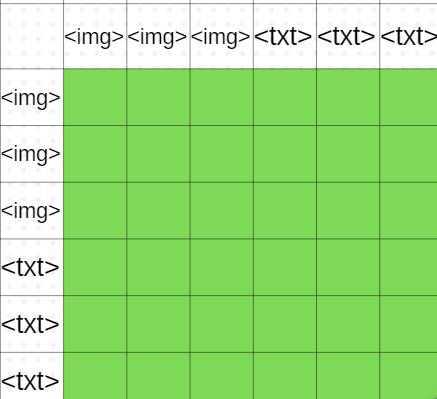}
        \caption{The attention mask for stage 0 training.}
        \label{fig:full_unmaskl}
    \end{subfigure}
    \hfill
    \begin{subfigure}[t]{0.3\textwidth}
        \centering
        \includegraphics[width=\linewidth]{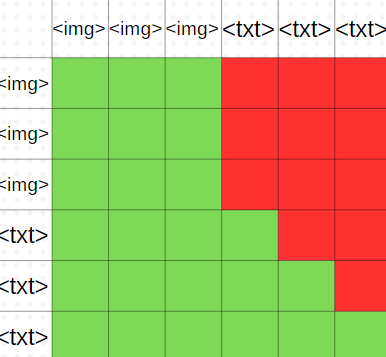}
        \caption{The attention mask for stage 1, 2, and 3 training.}
        \label{fig:attentionl}
    \end{subfigure}
    \caption{Visualization of attention masks across different training stages.}
    \label{fig:attention_masks_combined}
\end{figure}

\item \textbf{Stage-2}: Follows the same attention strategy as stage 1. Here the full model is trainable. We train the full model as there is not much degradation of language model capability after the introduction of image features, because of the alignment stages.
\item \textbf{Stage-3}: Follows the same attention strategy as stage 1, the full model is instructions-based fine-tuned on the LLaVA-Next dataset. This is instruction-based finetuning, where there is an instruction followed by the answer, the dataset has multiple instruction answers for a single image, and we consider them as a single conversation. The input data looks like this, 

\texttt{<im><im> <INST1> <ans1><ans1>  \\
                 <INST2> <ans2><ans2>}\\
Where \texttt{<im>} represents the image token, \texttt{<INST>} represents instructions given to the model, and \texttt{<ans>} represents the answer to the corresponding instruction.

Algorithm \ref{algo:2} presents the forward pass for stages 1, 2, and 3.
The model has been optimized on Cross-Entropy loss in all of the training stages and is given by:
\begin{equation}
L = - \frac{1}{M} \sum_{j=1}^{M} \sum_{i=1}^{N} y_{ij} \log(\hat{y}_{ij})
\end{equation}

Where \(N\) is the vocabulary size of the language model, \(M\) is the batch size, \(y_{ij}\) is the correct label, and \(\hat{y}_{ij}\) is the predicted probability for the \(i\)-th class of the \(j\)-th sample.

\begin{algorithm}
\caption{Algorithm for Model Forward Pass\\ for Stages 1, 2 \& 3}
\begin{algorithmic}[1]  
\State $V \gets \textsc{vision\_encoder}(image)$
\State $P \gets \textsc{projector}(V)$
\State $T \gets \textsc{text\_embeddings}(text)$
\State $L \gets \textsc{concatenate}(P, T)$
\State $attn\_mask \gets \textsc{update\_attention}(attn\_mask)$
\State $logits \gets \textsc{language\_model}(L, attn\_mask)$
\State $loss \gets \textsc{cross\_entropy\_loss}(logits, label\_text)$
\end{algorithmic}
\end{algorithm}
\label{algo:2}
\end{itemize}

\subsection{Training Methods}
Training transformer models are very memory and time consuming task. We apply many techniques to speed up the training and allowing models to compute bigger batch sizes. It is crucial to select a good batch size for better generalization of model. For training speedup we use faster attention SDPA \cite{Hosseini2024YouNT} attention, we use bfloat16 precision wherever possible and TF32 precision for all the operations that require 32-bit precision. To accommodate bigger batch-size we accumulates the gradient for required number of forward passes and average the gradients when we reach global batch size before backpropagation, we also do gradient checkpointing wherever required. Gradient clipping was done by normalizing the gradients for restricting model's weights movement by large amount.

\section{Performance Analysis}\label{sec4}

This section validates the proposed mechanism in terms of specifying the training platforms, and hyper-parameters.

\subsection{Training Platform and specification}

The experiments were conducted on an Ubuntu operating system using the PyTorch framework, with training performed on an NVIDIA A800 GPU with 80GB of memory. The training process for Stage 0 lasted for 1 hour, followed by Stage 1, also lasting 1 hour. In Stage 2, the model was initially trained for 4 hours, after which it underwent an additional 8 hours of training, bringing the total duration for this stage to 12 hours.

\subsection{Training Hyperparameters}
The model's vision Encoder has 400 million parameters, the language model has 500 million parameters, and the projector has 18 million parameters making the total parameters of the model 900 million as shown in table \ref{tab1}.

 The batch size for each of the stages is kept to 128. Each stage of training has a different learning rate, in stages 0 and 1, we keep the vision encoder and language model frozen so their learning rate is 0, as the projector has two linear layers we kept its learning rate to be 1e-3.
For stage 2 the learning rate for the vision encoder is 5e-6, the language model is 2e-5, and for the projector is 2e-3. For stage 3(fine-tuning) we only decreased the learning rate of the projector from 2e-3 to 1e-4 and the rest is kept the same as in stage 2. In each stage, the learning rate is decayed using Cosine decay to the minimum value of 1e-8 with AdamW optimizer with $\beta_1 = 0.9 \text{ and } \beta_2 = 0.999$ and weight decay of $0.01$. 
Different learning rates are mention in tables \ref{tab:table4}, \ref{tab:table5}, \ref{tab:table6} and \ref{tab:table7}.
The learning rate is decided based on the model's architecture and task. The transformer architecture converges the best with a learning rate of the order of 1e-5, and for linear layers, the suitable learning rate is of the order of 1e-3.

\begin{table}[htbp]
    \caption{Model Parameters}
    \centering
    \begin{tabular}{|c|c|c|c|}
        \hline
        \textbf{Vision Encoder} & \textbf{Projector} & \textbf{Language Model} & \textbf{Total} \\
        \hline
        400 Million & 18 Million & 500 Million & 900 Million \\
        \hline
    \end{tabular}
    \label{tab1}
\end{table}

\begin{table}[htbp]
    \centering
    \begin{subtable}[b]{0.45\linewidth}
        \centering
        \caption{Hyperparameters for Stage 0}
        \begin{tabular}{|c|c|}
        \hline
        \textbf{Hyperparameter} & \textbf{Value} \\ \hline
        Vision Encoder LR       & 0             \\ \hline
        Projector LR            & 1e-3          \\ \hline
        Language Model LR       & 0             \\ \hline
        Epochs                  & 1             \\ \hline
        \end{tabular}
        \label{tab:table4}
    \end{subtable}
    \hfill
    \begin{subtable}[b]{0.45\linewidth}
        \centering
        \caption{Hyperparameters for Stage 1}
        \begin{tabular}{|c|c|}
        \hline
        \textbf{Hyperparameter} & \textbf{Value} \\ \hline
        Vision Encoder LR       & 0             \\ \hline
        Projector LR            & 1e-3          \\ \hline
        Language Model LR       & 0             \\ \hline
        Epochs                  & 1             \\ \hline
        \end{tabular}
        \label{tab:table5}
    \end{subtable}
    
    \vspace{0.5cm} 

    \begin{subtable}[b]{0.45\linewidth}
        \centering
        \caption{Hyperparameters for Stage 2}
        \begin{tabular}{|c|c|}
        \hline
        \textbf{Hyperparameter} & \textbf{Value} \\ \hline
        Vision Encoder LR       & 5e-6          \\ \hline
        Projector LR            & 2e-3          \\ \hline
        Language Model LR       & 2e-5          \\ \hline
        Epochs                  & 1             \\ \hline
        \end{tabular}
        \label{tab:table6}
    \end{subtable}
    \hfill
    \begin{subtable}[b]{0.45\linewidth}
        \centering
        \caption{Hyperparameters for Stage 3}
        \begin{tabular}{|c|c|}
        \hline
        \textbf{Hyperparameter} & \textbf{Value} \\ \hline
        Vision Encoder LR       & 5e-6          \\ \hline
        Projector LR            & 1e-4          \\ \hline
        Language Model LR       & 2e-5          \\ \hline
        Epochs                  & 1             \\ \hline
        \end{tabular}
        \label{tab:table7}
    \end{subtable}
    
    \caption{Hyperparameters for different stages of training.}
    \label{tab:stage_comparison}
\end{table}

\begin{figure}[h]
    \centering
    \begin{subfigure}[b]{0.7\linewidth}
        \includegraphics[width=\linewidth]{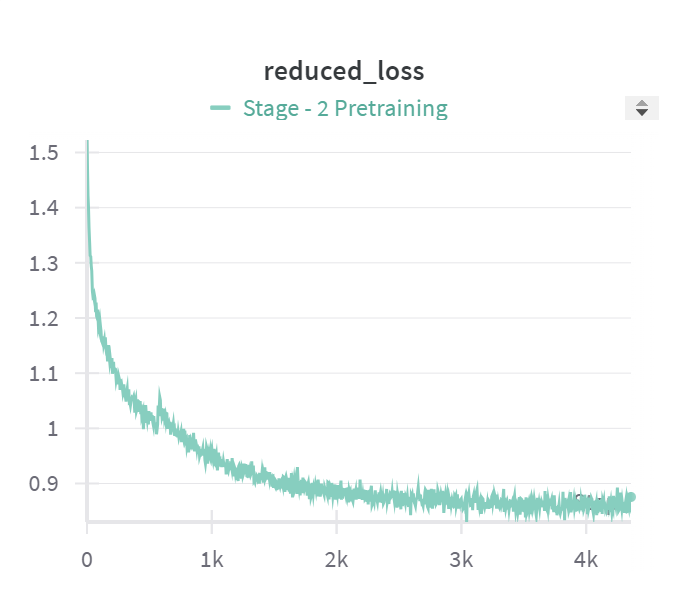}
        \caption{The training loss curve of Stage 2 Training.}
        \label{fig:stage2loss}
    \end{subfigure}
    \hfill
    \begin{subfigure}[b]{0.7\linewidth}
        \includegraphics[width=\linewidth]{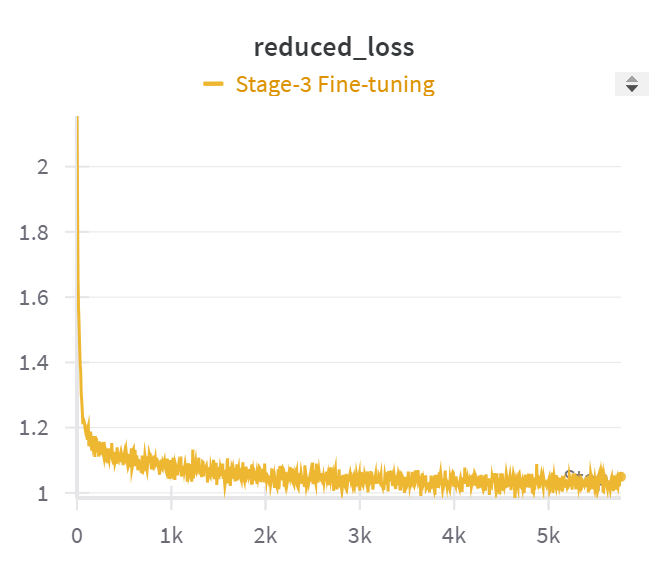}
        \caption{The training loss curve for Stage 3 Training.}
        \label{fig:stage3loss}
    \end{subfigure}
    
    \caption{Training loss curves for different stages of training.}
    \label{fig:stage_loss_comparison}
\end{figure}

\subsection{Results and Discussion}
This subsection details the evaluated results by measuring behavior and weight updation while improving the alignment. The training loss curves, refer to Fig.\ref{fig:stage2loss} and Fig. \ref{fig:stage3loss}, exhibit several notable characteristics, providing insights into the model's learning process:

\begin{itemize}

    \item \textbf{Asymptotic Behavior}: The loss curve follows a smooth, asymptotic decline which indicates effective learning. This gradual reduction in loss suggests that the model is optimizing the error function progressively over time. The asymptotic behavior demonstrates that the learning rate is well-tuned and the gradient updates are stable.

    \item \textbf{Convergence of Cross-Entropy Loss}: In the second stage of training, the cross-entropy loss converges to around a value of 0.85 as shown in Fig. \ref{fig:stage2loss} and around a value of 1.05 in stage-3 as shown in Fig. \ref{fig:stage3loss}. While this value indicates that the model has learned meaningful patterns from the data, the loss is not approaching zero. This could be due to factors such as class imbalance, inherent complexity in the data, or limitations in the model's capacity. This plateau suggests that the model might be reaching its optimal performance given the current configuration and dataset.

    \item  \textbf{Small Spikes with Each Weight Update}: The presence of small spikes in the loss curve, corresponding to each weight update represents minor adjustments made to the model's parameters at each step. Their relatively small amplitude suggests that the learning rate is appropriately set and tuned over time with cosine decay, allowing stable updates without causing significant oscillations or divergence. 
\end{itemize}

Overall, the loss curves suggest that the model is learning effectively, with stable updates and asymptotic convergence. However, the convergence at a cross-entropy loss of 0.8 implies potential room for improvement, which may be addressed by further tuning of hyperparameters, model architecture, or additional data preprocessing techniques.

\subsection{Model Performance}

The results presented in Table~\ref{tab:model_scores} provide a detailed comparison of model performance across three datasets: Flickr8k, Flickr30k, and COCO, using various evaluation metrics. The scores are reported for the Karpathy test splits, with different settings for full-shot and zero-shot scenarios.

The evaluation metrics include Bleu-1, Bleu-2, Bleu-3, and Bleu-4, which measure n-gram overlap between the predicted and reference captions. Higher scores indicate better performance. The \textit{METEOR} score takes into account synonymy and stemming, providing a more flexible measure of similarity. \textit{ROUGE-L} evaluates the longest common subsequence, assessing fluency and readability. Lastly, \textit{CIDEr} captures human consensus by computing consensus between the generated and reference sentences, with higher values reflecting better agreement.

Across the datasets, the model shows consistent performance, with higher scores on the COCO dataset. Specifically, the model achieves the highest Bleu-1 and CIDEr scores on COCO, which indicates its ability to generate captions that are more accurate and semantically meaningful on this dataset. Performance on Flickr8k and Flickr30k is slightly lower, especially in the zero-shot scenario, as expected due to the increased difficulty of generating captions for unseen data.

\begin{table}[htbp]
\centering
\caption{Model Scores on Flickr8k, Flickr30k, and COCO Karpathy Test-Splits}
\begin{tabular}{l|c|c|c}
\toprule
\textbf{Metric} & \textbf{Flickr8k} & \textbf{Flickr30k} & \textbf{COCO} \\

\midrule
Bleu\_1   & 76.0 & 74.1 & 79.2 \\
Bleu\_2   & 59.9 & 56.9 & 63.4 \\
Bleu\_3   & 45.3 & 42.2 & 48.6 \\
Bleu\_4   & 33.7 & 30.8 & 36.7 \\
METEOR    & 27.8 & 25.2 & 29.1 \\
ROUGE\_L  & 57.8 & 53.6 & 58.4 \\
CIDEr     & 99.4 & 80.5 & 127.1 \\

\bottomrule
\end{tabular}
\label{tab:model_scores}
\end{table}

\begin{table}[htbp]
\centering
\caption{Comparison of CIDEr Scores between Our Model and VILA on Flickr30k and COCO Datasets}
\begin{tabular}{l|c|c}
\toprule
\textbf{Model} & \textbf{Flickr30k (CIDEr)} & \textbf{COCO (CIDEr)} \\
\midrule
VILA           & 74.7                      & 118.5                  \\
GIT             & 49.6              & \textbf{144.8}         \\
Flamingo-80B    &67.2                 & - \\

BLIP-2 (13B)    &71.6                   & -\\
InstructBLIP(13B) &\textbf{82.8}                  & - \\

\textbf{Our Model(0.9B)}      & \textbf{*80.5}              & \textbf{*127.1}         \\
\bottomrule
\end{tabular}
\label{tab:comparison}
\footnotetext{The best and second best are represented by \textbf{best} and \textbf{*second best} respectively.}
\end{table}

The model achieves a higher CIDEr score than VILA \cite{Lin2023VILAOP}, on COCO and Flcikr30k datasets. Our model scored 25 points higher in Flick8k and 6 points higher in Flickr30k when compared to a much bigger model (0.9 billion vs 7 and 13 billion). On the COCO caption dataset, our model achieves 10 points more than VILA. When compared to  GIT2 \cite{Wang2022GITAG}, on the Flickr30k dataset our model performs better on 0-shot evaluation but shows worse performance on the COCO dataset. Our model performs better than popular models like Flamingo \cite{Alayrac2022FlamingoAV}, our 0.9 billion parameter model achieves 13 CIDEr points higher on FLickr30k, and is just 2 CIDEr scores behind InstructBLIP-13 billion model, even though our model was trained on 1\% of the data size of InstructBLIP. The model even surpassed the BLIP2-13 billion model, this shows our model's adaptive capabilities to various datasets. 

\subsection{Qualitative Analysis}
Below are the generated text from the our model on various datasets.

\begin{table}[htbp]
\centering
\caption{Visual Question Answering for flickr8k.}

\begin{tabular}{|c|c|}
\hline
\includegraphics[width=0.15\textwidth]{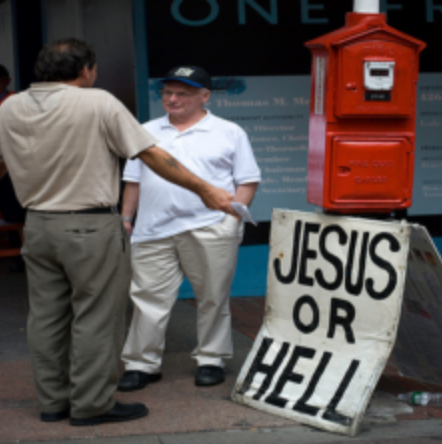} & 
\begin{tabular}{@{}l@{}}
\textbf{ques1}: how many men are there? \\
\textbf{Ans}: 2
\\ \\ 
\textbf{ques2}: What is the man on the right \\wearing?\\
\textbf{Ans}: The man on the right is wearing \\ a hat.
\\ \\
\textbf{ques3}: Where is the poster? \\
\textbf{Ans}: On sidewalk
\\ \\
\textbf{ques4}: What does the sign say?\\
\textbf{Ans}: Jesus or hell \\ \\

\end{tabular} \\
\hline
\includegraphics[width=0.15\textwidth]{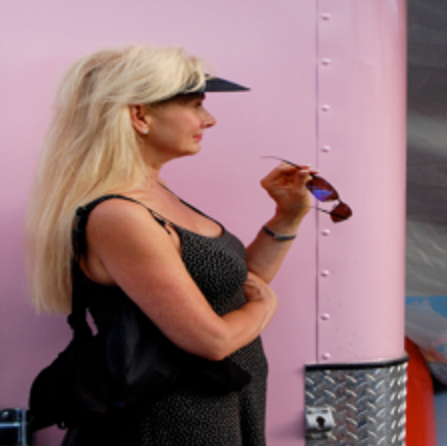} &
\begin{tabular}{@{}l@{}}
\textbf{Ques1}: What is that woman holding? \\
\textbf{Ans}: Sunglasses\\ \\

\textbf{Ques2}: What is the color of background?\\
\textbf{Ans}: Pink\\ \\ 

\textbf{Ques3}: What color is woman's hair? \\
\textbf{Ans}: Blond\\\\\\
\end{tabular} \\
\hline

\includegraphics[width=0.15\textwidth]{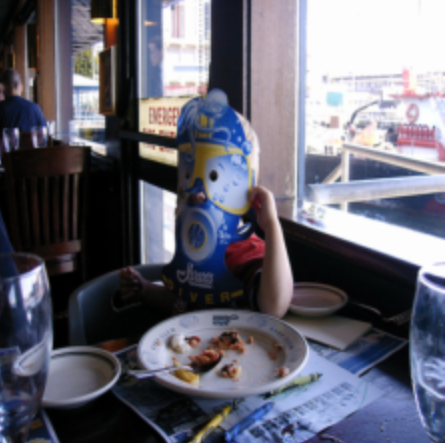} &
\begin{tabular}{@{}l@{}}
\textbf{Ques1}: Explain the image in details \\ \\ 
\textbf{Ans:} In the image, a young boy is sitting at \\ a table in a restaurant, holding a blue and \\ yellow stuffed animal. He is eating a meal, \\ which includes a plate of food and a cup of \\ coffee. The boy is also holding a fork and \\ a knife, suggesting that he is in the process \\ of eating. The table is set with a few other \\ items, including a cup, a plate, and a bowl, 
 \\ indicating that the meal is ready to be \\ enjoyed. \\\\
\end{tabular} \\
\hline
\end{tabular}
\label{tab: vqa}
\footnotetext{Proposed Model's response on visual question answering task.}
\end{table}




1) \textbf{Visual question Answering}
The fine-tuning of the model is done on visual instruction dataset, LLaVA-Next-790k, The results on visual-question answering task is given in table \ref{tab: vqa}.
2) \textbf{Qualitative results on Healthcare dataset}\\
This adaptive model is trained on the PathVQA dataset \cite{He2020PathVQA3Q}, which includes 19k image-question-answer triplets. The model demonstrates promising results, showcasing its potential in handling visual question answering tasks within medical contexts. The results on visual question-answering on PathVQA dataset \cite{He2020PathVQA3Q} are shown in table \ref{tab: pathvqa}.

\begin{table}[htbp]
\centering
\caption{Visual Question Answering on PathVQA Dataset.}

\begin{tabular}{|c|c|}
\hline
\includegraphics[width=0.20\textwidth]{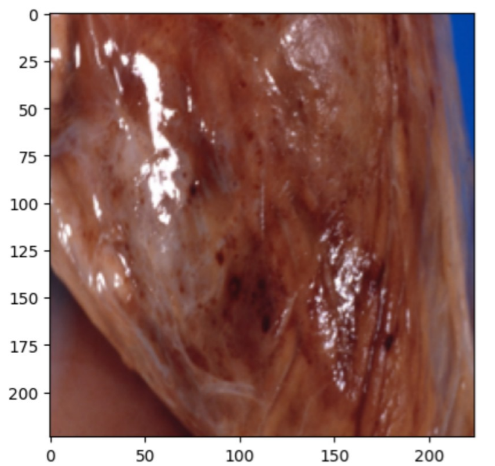} & 
\begin{tabular}{@{}l@{}}
\textbf{Instruction:} is muscle present? \\ \textbf{Original Answer: } yes \\ \textbf{Predicted Answer:} yes \\
\end{tabular} \\
\hline
\includegraphics[width=0.20\textwidth]{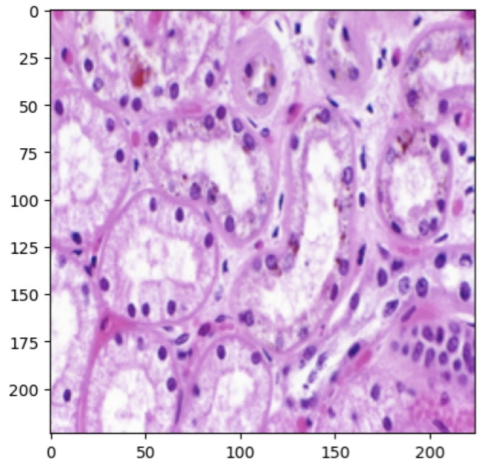} &
\begin{tabular}{@{}l@{}}
\textbf{Instruction:} where is this? \\ \textbf{Original Answer: } urinary \\ \textbf{Predicted Answer:} urinary \\
\end{tabular} \\
\hline

\includegraphics[width=0.20\textwidth]{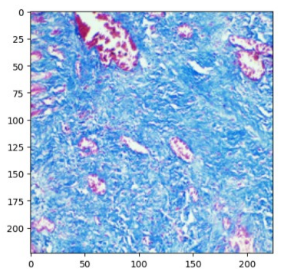} &
\begin{tabular}{@{}l@{}}
\textbf{Instruction:} What is stained blue\\ by the trichrome stain?  \\ \textbf{Original Answer: } Collagen \\ \textbf{Predicted Answer:} The collagen \\
\end{tabular} \\
\hline
\end{tabular}
\footnotetext{Proposed Model's response on medical question answering task after finetuning on PathVQA.}
\label{tab: pathvqa}
\end{table}

\subsection{Other Findings}

\begin{figure}[h]
    \centering
    \begin{subfigure}[b]{0.6\linewidth}
        \includegraphics[width=\linewidth]{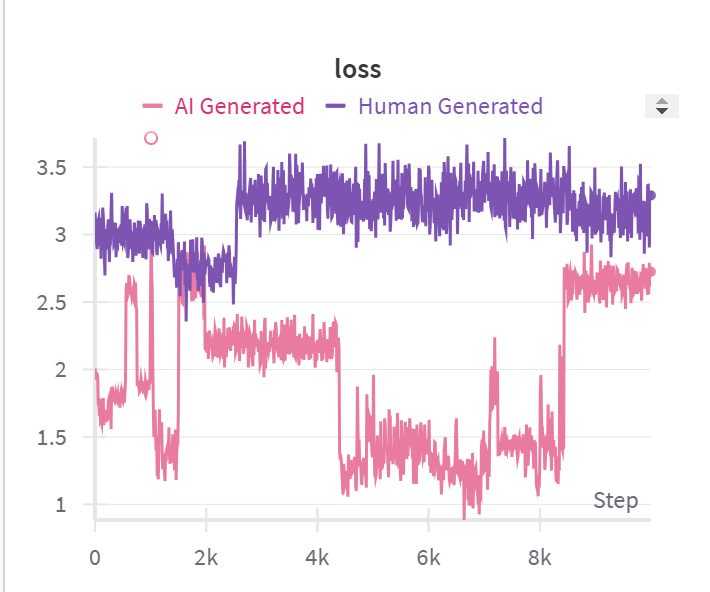}
        \caption{The training loss comparison for AI-Text data and Human Text data.}
        \label{fig:other_finding}
    \end{subfigure}
    \hfill
    \begin{subfigure}[b]{0.6\linewidth}
        \includegraphics[width=\linewidth]{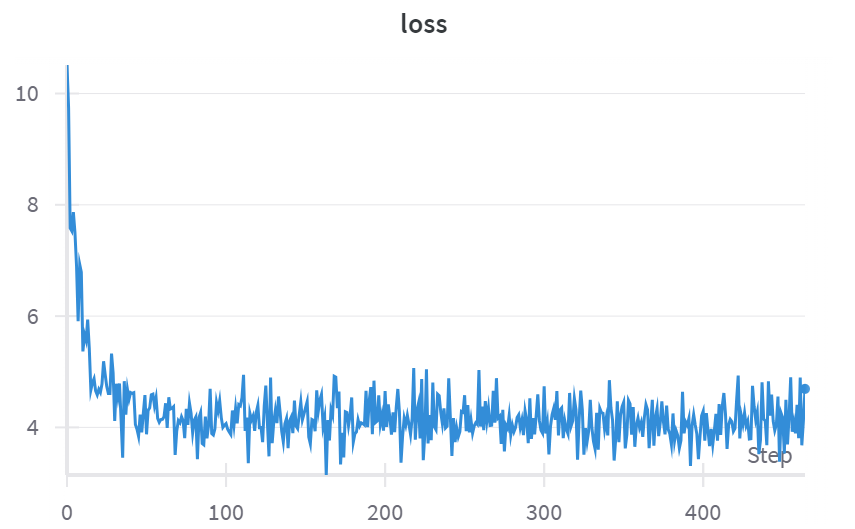}
        \caption{The training loss curve for Training on Human Captionized data.}
        \label{fig:bad_loss}
    \end{subfigure}
    
    \caption{Training loss comparisons for different datasets.}
    \label{fig:loss_comparison}
\end{figure}

In our experiments, we observed an interesting behavior when comparing model convergence on two datasets that shared the same images but differed in the source of captions. One dataset utilized human-generated captions, while the other employed AI-generated captions. Remarkably, the model demonstrated faster convergence when trained with AI-generated captions. The loss for human-captioned data is shown in Fig. \ref{fig:bad_loss}. It can be seen, that the model converges to a loss of around 5, which is quite high for the text generation objective.

Moreover, a comparative analysis of the cross-entropy loss between the two types of texts revealed that AI-generated texts consistently resulted in lower loss values for semantically similar inputs. This suggests that AI-generated text, while potentially less nuanced than human-generated text, may align more efficiently with the model’s learning process, especially in terms of semantic consistency and simplicity. The loss comparison between AI and human-generated text is shown in Fig. \ref{fig:other_finding}, the loss is a cross-entropy loss for the next token prediction with the same Qwen2.5, 0.5 billion model. It can be seen that the loss of AI-generated text (pink) is lesser than human-generated (purple) text. These findings could have significant implications for the use of AI-generated data in training models, particularly when faster convergence and reduced loss are critical factors.

\section{Limitation and Future Scope of the paper}\label{sec5}
\begin{figure}[h]
    \centering
    \begin{subfigure}[b]{0.45\linewidth}
        \includegraphics[width=\linewidth]{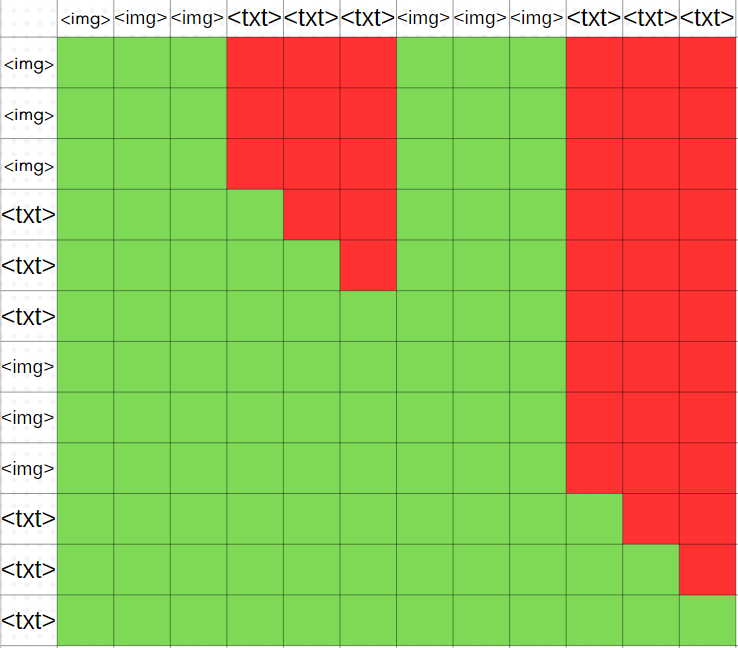}
        \caption{The attention mask for interleaved datasets (Figure 1).}
        \label{fig:future_work}
    \end{subfigure}
    \hfill
    \begin{subfigure}[b]{0.45\linewidth}
        \includegraphics[width=\linewidth]{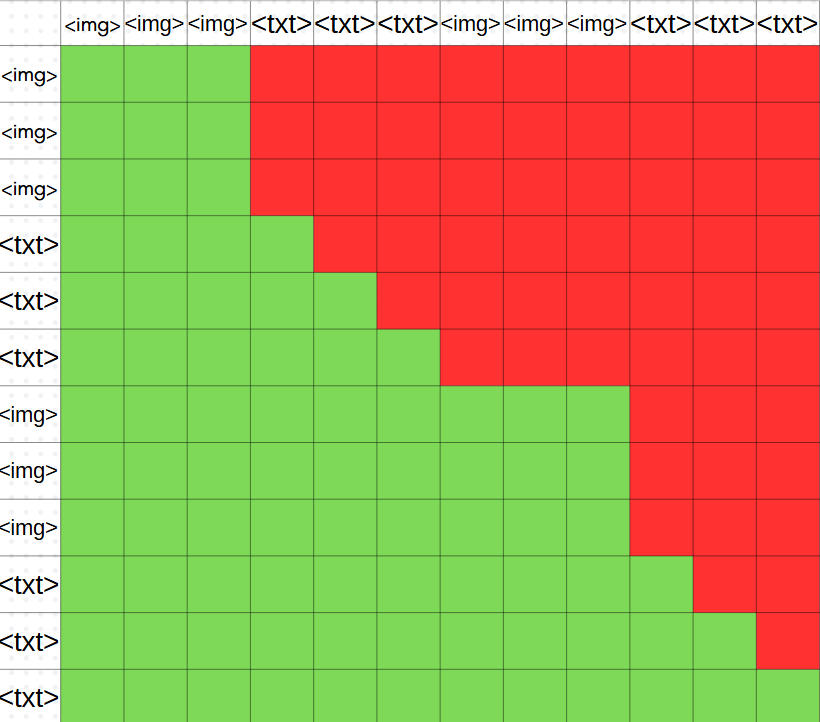}
        \caption{The attention mask for interleaved datasets (Figure 2).}
        \label{fig:future_work2}
    \end{subfigure}
    \caption{Visualization of attention masks for interleaved datasets.}
    \label{fig:future_work_combined}
\end{figure}

With the rapid development of vision language models, new formats of data for training these models are also a hot topic of research in this field. Flamingo\cite{Alayrac2022FlamingoAV} introduced an interleaved image-text dataset, and an open-source interleaved dataset has been introduced, MMC4\cite{Zhu2023MultimodalCA}. The data is represented as follows: 
\\
\texttt{<im1><text> <im2><txt2> <im3><txt3>>}
\\
Images and text are considered as a part of a long sequence, this sequence contains multiple images and test descriptions, unlike our model where in a sequence only one image-text pair is present. For training on interleaved datasets, a very large context-length is required which is a very computation-consuming task. Following our work of attention masks on a single pair of image-pair, a similar masking technique can be used for training on interleaved datasets, the attention mask is shown in Fig. \ref{fig:future_work}. Different variations can be experimented with, for example in Fig. \ref{fig:future_work}, the images and text can pay attention to future image tokens but not to future text tokens, and in Fig. \ref{fig:future_work2}, the images and text cannot pay attention to future images and text.

\section*{Conclusion and Future Directions}\label{sec6}
In the proposed work, we explored different architectures for vision-language models, the importance of choosing correct attention masking strategies, the importance of alignment of the modalities before training the language model (unfreezing the weights), and the choice of the correct dataset for efficient and faster convergence of models. The gradual learning from projector to language model is crucial as it doesn't degrade the performance of language model. After all the training stages the proposed model was able to perform well in various vision-language task like image captioning and visual question answering, further, it's performance in medical visual question answering shows the good adaptability of the model. We hope these findings will help in the further development of attention-based multimodal models.

\bibliographystyle{IEEEtran}
\bibliography{IEEEabrv, citation}

\end{document}